\documentclass[letterpaper, 10 pt, conference]{ieeeconf}
\IEEEoverridecommandlockouts
\usepackage{cite}
\usepackage[table]{xcolor}
\usepackage{amsmath,amssymb,amsfonts}
\usepackage{algorithmic}
\usepackage{graphicx}
\usepackage{textcomp}
\usepackage{xcolor}
\usepackage{booktabs}
\usepackage{makecell}
\usepackage{pifont}
\usepackage[dvipsnames]{xcolor}
\usepackage{bm}

\def\BibTeX{{\rm B\kern-.05em{\sc i\kern-.025em b}\kern-.08em
    T\kern-.1667em\lower.7ex\hbox{E}\kern-.125emX}}
\begin{document}

\title{\LARGE \bf HiST-VLA: A Hierarchical Spatio-Temporal Vision-Language-Action Model for End-to-End Autonomous Driving\\
}
\author{
    Yiru Wang$^{1}$,
    Zichong Gu$^{2}$,
    Yu Gao$^{1}$, 
    Anqing Jiang$^{1}$,
    Zhigang Sun$^{1}$,
    Shuo Wang$^{1}$, 
    Yuwen Heng$^{1}$, 
    Hao Sun$^{1}$
    \thanks{
    This work was supported by Bosch Corporate Research. (\textit{Corresponding author: : hao.sun4@cn.bosch.com})} 
    \thanks{$^{1}$
            Yiru Wang,
            Yu Gao,
            Anqing Jiang,
            Zhigang Sun,
            Shuo Wang,
            Yuwen Heng,
            Hao Sun are with Bosch Corporate Research, Bosch (China) Investment Ltd., Shanghai, China. }
    \thanks{$^{2}$
        Zichong Gu
        is with School of Communication and Information Engineering, Shanghai University, Shanghai, China
    }
}

\maketitle

\begin{abstract}
Vision-Language-Action (VLA) models offer promising capabilities for autonomous driving through multimodal understanding. However, their utilization in safety-critical scenarios is constrained by inherent limitations, including imprecise numerical reasoning, weak 3D spatial awareness, and high sensitivity to context. To address these challenges, we propose HiST-VLA, a novel \textit{\textbf{Hi}}erarchical \textit{\textbf{S}}patio-\textit{\textbf{T}}emporal VLA model designed for reliable trajectory generation.

Our framework enhances 3D spatial and temporal reasoning by integrating geometric awareness with fine-grained driving commands and state history prompting. To ensure computational efficiency, we integrate dynamic token sparsification into the VLA architecture. This approach fuses redundant tokens rather than filtering them, effectively reducing redundancy without sacrificing model performance. Furthermore, we employ a hierarchical transformer-based planner to progressively refine coarse VLA waypoints into fine-grained trajectories. Crucially, the planner utilizes dynamic latent regularization to incorporate language commands, ensuring strict spatial grounding and temporal coherence. Extensive evaluation on the NAVSIM v2 benchmark demonstrates state-of-the-art performance on Navtest, achieving an EPDMS of 88.6, and EPDMS of 50.9 on pseudo closed-loop Navhard benchmark. 
\end{abstract}
\section{Introduction}
Contemporary autonomous driving (AD) systems operate through a cascaded pipeline of perception, prediction, planning, and control to achieve real-time environment understanding and behavior generation. While demonstrating reliability in typical scenarios, these architectures remain susceptible to failures at module boundaries and often underperform in complex social interactions or require deeper semantic reasoning.

Recent advances in vision-language models (VLMs) have shown extensive semantic knowledge and strong generalization to unfamiliar inputs. They can interpret complex scenes and detect rare objects often missed by conventional perception modules. Integrating these capabilities into AD systems by fine-tuning pre-trained VLMs on diverse driving datasets, can significantly enhance robustness in challenging scenarios and enable human-AD interaction. A prevalent strategy for such integration is VLM-enhanced end-to-end (E2E) dual-system architecture~\cite{jiang2025diffvla, jiang2024senna, jiang2025irl}. However, a key challenge lies in the loose coupling between high-level language outputs and low-level vehicle control, which can lead to unreliable reactions to natural instructions and increased false positives. Additionally, performance heavily depends on large-scale annotated datasets, while intermediate representations often impair latency and generalizability. Thus, despite improving interpretability, loosely coupled VLM systems remain insufficient for generating fully trustworthy driving actions.

Other than VLM dual systems, recent Vision-Language-Action (VLA) models can be broadly categorized into two architectural paradigms: E2E VLA trajectory prediction models, and hybrid approaches where the VLA generates coarse trajectories that are subsequently refined by a dedicated trajectory head. Among these, E2E VLAs~\cite{arai2025covla, hwang2024emma, renz2025simlingo, zhou2025opendrivevla}, which directly map multimodal inputs to trajectory sequences, can face limitations in aspects such as spatial reasoning, 3D geometric interpretation, and action-space granularity, affecting the model's ability of detailed situational understanding and precise trajectory generation, especially in highly complex or unstructured driving scenarios.

In contrast, the hybrid architecture~\cite{li2025recogdrive, fu2025orion, tian2024drivevlm}, where a VLA generates an initial trajectory or trajectory tokens, subsequently refined by a secondary trajectory optimization head, offers a balance between architectural consistency and functional performance. This approach maintains the VLA as the core planner while compensating for its perceptual and representational limitations through a dedicated refinement module. The trajectory head, often implemented as a lightweight network or an optimization layer, enhances the physical plausibility, smoothness, kinematic feasibility, and task compliance of the resulting trajectory. This method is increasingly regarded as a ``pure” VLA model extension, as it builds upon and refines the VLA’s inherent capabilities without relying on additional E2E systems.

To address key limitations in current VLAs, such as the lack of granular 3D spatial awareness, insufficient temporal coherence, and difficulties in generating precise control-ready trajectories, we propose HiST-VLA: a \textit{\textbf{Hi}}erarchical \textit{\textbf{S}}patially-grounded and \textit{\textbf{T}}emporally coherent VLA that integrates 3D geometric reasoning and temporal modeling with multi-stage trajectory refinement. Unlike prior hybrid designs, our model introduces a novel hierarchical architecture supporting multi-stage decoding of coarse VLA trajectories, which enhances trajectory safety, comfort, and interpretability without auxiliary E2E models or costly annotations.

The spatio-temporal VLA structure integrates enhanced 3D spatial understanding through explicit geometric encoding and temporal state prompting to capture historical states. This module further incorporates a dynamic token sparsification mechanism based on self-similarity in attention space, effectively reducing redundancy while preserving semantically critical features. Within an extended chain-of-thought (CoT) reasoning process, fine-grained driving commands are interpreted to strengthen spatial and behavioral alignment. The model outputs trajectories accompanied by confidence scores, explicitly indicating prediction certainty. This design not only enhances spatio-temporal grounding but also enables efficient, semantics-aware reasoning, supporting the generation of precise and reliability-conscious trajectories.

Furthermore, we propose a hierarchical planner that refines the coarse trajectory from the VLA through structured processing stages. The pipeline employs a variational autoencoder (VAE) to learn compact latent representations of feasible trajectories, further enabling efficient semantic alignment with interpreted driving commands. This is followed by a scorer-guided optimization step to enhance safety, feasibility, and comfort. This multi-granular refinement strategy significantly extends conventional single-stage decoders, effectively bridging high-level instructions and executable controls in a fully differentiable architecture.

To ensure a comprehensive and realistic validation of our VLA-based driving system, we leverage the NAVSIM v2 benchmark~\cite{dauner2024navsim}. This framework provides a rigorous evaluation that integrates both open-loop and pseudo closed-loop testing, offering a more reliable and robust assessment of real-world driving performance compared to traditional offline metrics.

Our contributions can be summarized as follows:
\begin{itemize}
\item HiST-VLA Architecture: A novel framework that integrates spatial grounding and temporal coherence with multi-stage trajectory refinement in a fully integrated framework, enhancing trajectory comfort and safety.

\item Efficient Spatio-Temporal Representation with Token Sparsification: Incorporates 3D encoding and temporal state prompting with a dynamic token sparsification mechanism. By leveraging self-similarity guided token fusion, our approach effectively reduces redundancy while preserving semantically critical features.

\item Semantics-Aligned Hierarchical Planning: Refines trajectories via a multi-stage pipeline aligning granular driving commands with motion, ensuring safety and comfort trajectory generation through confidence-aware regularization and multi-criteria scoring.
\end{itemize}
\vspace{-5pt}
\section{Related Work}
\subsection{E2E Autonomous Driving}
\vspace{-2pt}
Autonomous driving systems have transitioned to E2E approaches that map sensors directly to controllable waypoints~\cite{chitta2022transfuser, hawke2020urban}. To address challenges in data efficiency and robustness, recent works introduce efficient vectorized (VAD~\cite{jiang2023vad}) and sparse architectures (SparseDrive~\cite{sun2024sparsedrive}). While optimization-based refinement exists~\cite{chen2024vadv2}, newer methods focus on expressive policies via iterative refinement (iPad~\cite{guo2025ipad}). SimScale further improves generalization through scalable simulation and data co-training~\cite{tian2025simscale}. Despite these advances, most E2E methods still lack semantic interpretability and natural language interfaces for human-vehicle interaction. 
\subsection{VLM for E2E Autonomous Driving}
The integration of linguistic modalities leverages LLMs and VLMs to enhance reasoning, interpretability, and generalization in vehicle decision-making. By leveraging large language models (LLMs) and VLMs~\cite{touvron2023llama, liu2023visual}, modern systems unify perception and reasoning within a shared semantic space, incorporating commonsense knowledge (e.g., ``flashing hazard lights" implies ``vehicle in distress") to improve adaptation to novel scenarios~\cite{jiang2025diffvla, jiang2024senna}. 

However, foundation models face challenges including limited spatial awareness, imprecise outputs, high latency, and hallucination risks. Recent work addresses these limitations by improving spatial representations~\cite{tian2024tokenize, winter2025bevdriver, zhai2025world} and reduce latency through distillation or advisory architectures~\cite{chen2024asynchronous, pan2024vlp}. In spite of these advancements, current VLM-based driving systems often lack closed-loop control integration, and struggle to align semantic reasoning with safe, continuous motion planning. 
\begin{figure*}[htbp!] %
    \centering
    \includegraphics[width=7.0in]{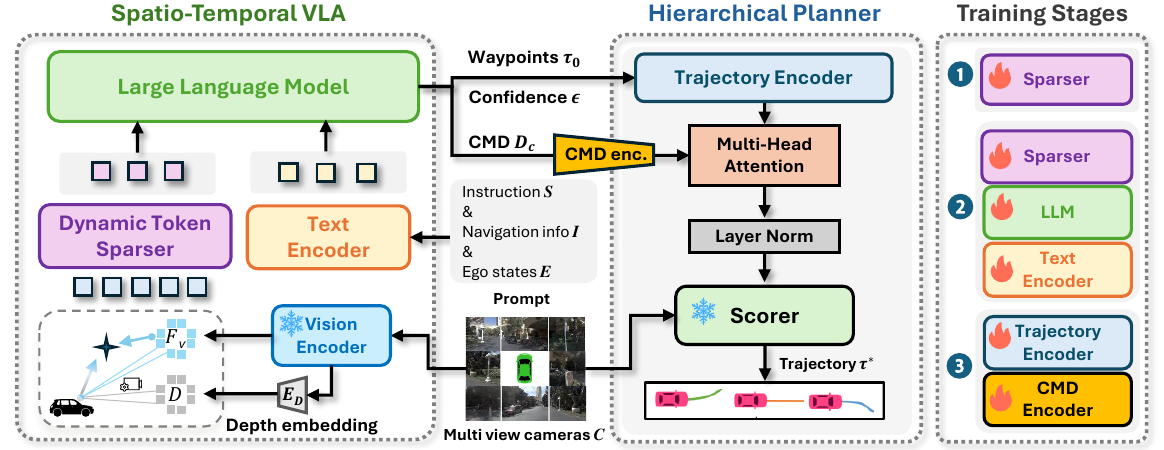} %
    \vspace{-18pt}
    \caption{The Framework of the proposed HiST-VLA, including the Spatio-Temporal VLA architecture, hierarchical planner and main training stages.}
    \vspace{-8pt}
    \label{fig:pipeline} %
\end{figure*}
\subsection{VLA in E2E Autonomous Driving}
Inspired by advances in embodied intelligence~\cite{black2024pi_0, kim2024openvla}, recent works integrate multimodal perception and language into unified architectures to bridge the semantic-control gap. These VLA models forms cohesive perception-reasoning-control pipelines that support adaptation to rare scenarios, robustness under partial observability, and comprehension of high-level instructions~\cite{xu2024drivegpt4, tian2024drivevlm}. Built upon VLMs, VLAs exhibit strong cross-domain generalization. Some explicitly model multi-step CoT reasoning for interpretability, as in DriveCoT~\cite{wang2024drivecot}. Others focus on aligning language to actions: Orion bridges a language reasoning space and a trajectory action space via a generative planner~\cite{fu2025orion}, while SimLingo unifies closed-loop driving with vision-language understanding and language-action alignment~\cite{renz2025simlingo}. To improve driving quality, others incorporate advanced optimization mechanisms: RecogDrive utilizes a diffusion-based planner to generate diverse, human-like trajectories~\cite{li2025recogdrive}, while IRL-VLA employs a reward world model to enable efficient closed-loop optimization without relying on high-fidelity simulation~\cite{jiang2025irl}.

Despite these advances, current VLAs often lack the granular 3D spatial reasoning and temporal coherence required for robust, control-ready planning in dynamic environments. 
\section{Method}
In this paper, we propose HiST-VLA, an E2E framework that integrates perception, semantic reasoning, and trajectory planning within a unified architecture. The framework comprises: (1) a unified architecture combining 3D-aware geometric representation and semantic reasoning for multi-stage trajectory refinement, (2) a Spatio-Temporal VLA backbone featuring multi-view 3D understanding, temporal modeling, CoT reasoning and dynamic token sparsification to reduce redundancy; and (3) a hierarchical planner that generates continuous trajectories from coarse VLA outputs via semantic conditioning and multi-criteria scoring. 
\vspace{-2pt}
\subsection{HiST-VLA Framework}
\vspace{-2pt}
\label{subsec:HiST_VLA_Framework}
As illustrated in Fig.~\ref{fig:pipeline}, the Spatio-Temporal VLA and hierarchical planner enable spatial grounding, temporal modeling and precise trajectory refinement through a multi-stage process. The Spatio-Temporal VLA acts as the primary policy generator, responsible for producing an initial driving plan and coarse trajectories. Given multi-view camera inputs $\mathcal{C}$ of the current timestep $t$, the Spatio-Temporal VLA first processes the visual inputs through a vision encoder to obtain $F_v$. This representation is further enhanced with depth estimation $\bm{E_D}$ and the combined 3D visual features are processed by a dynamic token sparser to produce spatially-aware visual tokens. Concurrently, the instruction $S$, the time-series navigation information $I = (i_t, i_{t-1}, ..., i_{t-k})$ and time-series ego-vehicle states $E = (e_t, e_{t-1}, ..., e_{t-k})$ for the past $k$ frames are encoded by a text encoder. These visual tokens and textual embeddings are then fed into the large language model (LLM). Following a CoT reasoning process, the model then auto-regressively generates three outputs based on the derived reasoning steps: a granular driving command $D_c$ for the next four seconds, a candidate trajectory $\tau_0$ with a confidence score $\epsilon$.

Then the hierarchical planner optimizes the coarse trajectory $\tau_0$. It first encodes $\tau_0$ into a latent space using a VAE and then semantically aligns it with the granular driving command $D_c$. A multi-criteria scorer then assesses feasibility, safety, and comfort, refining it through additive offsets to produce an optimized trajectory ${\tau^*}$. This multi-stage refinement mechanism forms the core of our hierarchical planning strategy, enabling progressive refinement from coarse trajectory and semantic intention to precise motion. 
\vspace{-2pt}
\subsection{Spatio-Temporal VLA Model}
\vspace{-2pt}
\label{subsec:Spatio_Temporal_vla}
\begin{figure*}[htbp!]
\centerline{\includegraphics[width=7.2in]{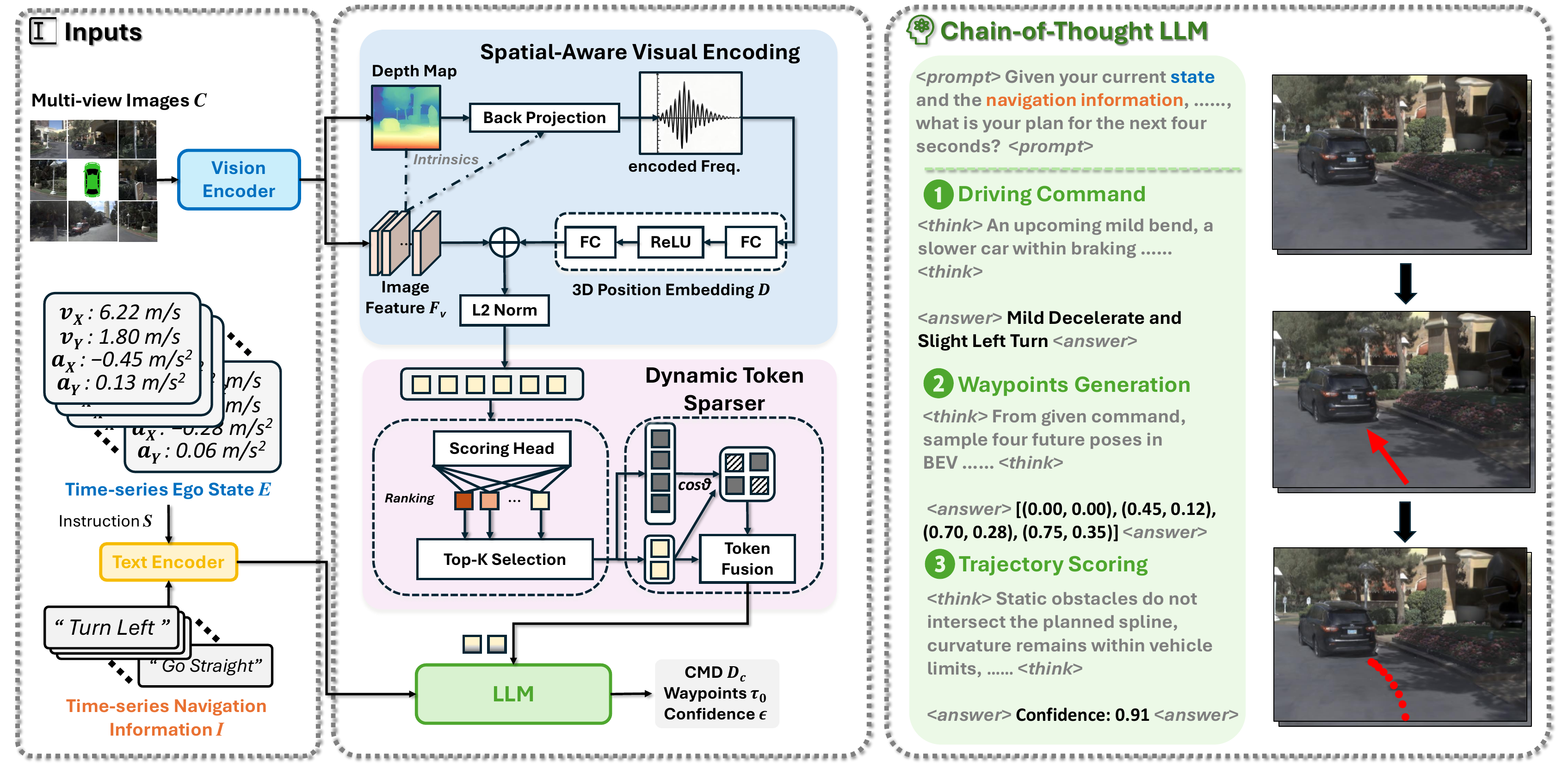}}
\vspace{-12pt}
\caption{Spatio-Temporal VLA Model: Leveraging multi-view images,  long-term navigation information, and ego state, our model employs CoT reasoning and dynamic token sparsification to produce a granular driving command and a coarse trajectory with confidence for the next four seconds.}
\vspace{-12pt}
\label{fig:Spatio_Temporal_vla}
\end{figure*}
Fig.~\ref{fig:Spatio_Temporal_vla} illustrates the Spatio-Temporal VLA model, which integrates spatially-aware visual inputs and temporal history to enable structured reasoning. Its architecture is detailed in the following sub-sections. 

\subsubsection{Spatially-Aware Visual Encoding}
\label{subsec:Spatially_Aware_Visual_Encoding}
We extract 2D patch features from multi-view images using a ViT-L/14 from CLIP~\cite{radford2021learning} as vision encoder. To recover 3D scene geometry, a pretrained monocular depth estimator~\cite{bhat2023zoedepth} generates dense depth maps. Using camera intrinsics, each 2D image patch is back-projected into the ego-vehicle's 3D coordinate frame. The resulting 3D coordinates are encoded via a Multi-Layer Perceptron (MLP) with positional embeddings, and this encoded 3D information is added to the original 2D features. This process yields spatially-augmented visual tokens that explicitly encode the metric structure of the scene.
\vspace{-2pt}
\subsubsection{Dynamic Token Sparser} 
\label{subsec:dynamic_token_sparser}
Unlike conventional methods that apply MLPs to transform all image tokens equally without explicit selection~\cite{liu2024improved, jiang2024senna}, our approach leverages the intrinsic self-attention scores of image tokens to perform adaptive token sparsification. This module is built upon the  intra-block Joint Token Pruning and Squeezing (eTPS)~\cite{wei2023joint}. By retaining tokens with high self-attention activation and merging less salient ones, we achieve more efficient and feature-aware visual token reduction. The top-$k$ most informative tokens are retained, while the remaining tokens are adaptively merged into the kept ones through a learnable transformer-based fusion module. This self-similarity guided merging mechanism preserves structurally critical regions while reducing computational sequence length. The resulting compressed visual tokens are projected into the LLM’s embedding space to enable efficient multimodal interaction.
\vspace{-5pt}
\subsubsection{Temporal State Modeling} 
Furthermore, to enhance the smoothness and comfort of the generated driving commands and trajectories, we incorporate temporal modeling into the text prompt. Specifically, the long-term navigation information $I$ (discrete maneuver labels: left, straight, right \cite{dauner2024navsim}) and ego vehicle states $E$ (longitudinal/latitudinal speed/velocity: $v_x$, $v_y$, $a_x$, $a_y$) for the previous $k$ frames are used as part of the prompt for the current step $t$. The above text, along with the user instruction (\textit{What are your predicted driving intention and waypoints for next four seconds?}), is inserted into the prompt. This prompt conditioning transforms the model from a reactive function $\pi(a_t | s_t)$ to a more informed function $\pi(a_t | s_t, s_{t-1}, ..., s_{t-k})$, where $s_t = (i_t, e_t)$, improving temporal coherence, reducing abrupt command changes, and yielding more continuous, human-like trajectory planning that better aligns with comfort expectations.
\vspace{-2pt}
\subsubsection{Granular Meta-Action and Confidence-Aware Trajectory Reasoning Enhanced CoT} %
\label{subsec:cot}
The CoT performs structured reasoning to generate justified driving actions. Given the fused multimodal input, the LLM first analyzes the ego state and perceived scene context, interprets the high-level semantic intent conveyed by the language instruction (e.g., ``Turn Left"), and proposes a granular driving command for the next four seconds. Prior work in meta-action planning often struggles to capture the nuanced intent embedded in language instructions, relying on simplistic categorical actions~\cite{jiang2025diffvla, jiang2024senna, jiang2025alphadrive} (e.g., Lateral: Left, Straight, Right; Longitudinal: Accelerate, Keep, Decelerate), which lack the expressiveness for nuanced vehicle control. Such coarse discretizations fail to capture essential behavioral variations, resulting in suboptimal performance in complex scenarios.

To overcome this and further enhance the spatial awareness of VLA, we introduce a finely-grained meta-action schema designed directly from the geometric and kinematic characteristics of real-world trajectories. Rather than relying on basic directional and speed labels, our method derives action distinctions from the shape and profile of future trajectory truths. For lateral control, we propose a diverse set of primitives including: Sharp\_Left\_Turn, Slight\_Left\_Turn, Left\_LaneChange, Sharp\_Right\_Turn, Slight\_Right\_Turn, Right\_LaneChange, Straight\_Strict, and Lane\_Micro\_Adjust. This reflects tangible differences in curvature and lateral displacement, allowing the model to differentiate between subtle variations of a single language command like “Turn Left”.

For longitudinal control, we define a hierarchy of acceleration profiles based on kinematic patterns: Full\_Stop, Creeping, Emergency\_Decel, Mild\_Decel, Controlled\_Decel, Constant\_Speed\_Loose, Constant\_Speed\_Strict, Mild\_Accel and Aggressive\_Accel, covering a spectrum from precise speed keeping to safety-critical deceleration. This granularity allows the model to precisely execute complex instructions that involve speed adjustments.

The proposed granular driving command reasoning, translating semantic intent into precise actions, significantly improves discriminative capacity for decision-making and generates smoother, more human-like behavior. Based on this command, the model generates a coarse trajectory $\tau_0$ that geometrically instantiates the intended maneuver. Finally, the model estimates a confidence score $\epsilon$ for the proposed trajectory. This score is derived derived from token confidences. For each token, the final transformer layer of the VLA produces a probability distribution via a softmax function, and the token’s confidence is the maximum probability in this distribution. The overall trajectory confidence $\epsilon$ is then calculated as the geometric mean of these individual token confidences. This multi-step reasoning approach not only decomposes the decision process into human-interpretable stages but also contributes to improved trustworthiness in motion planning.
\vspace{-2pt}
\subsection{Transformer-based Hierarchical Planner}
\vspace{-2pt}
\label{subsec:hierarchical_planner} 
The coarse trajectory $\tau_0$ from the Spatio-Temporal VLA serves as a semantic and structural prior for the hierarchical planner. In the first refinement stage, we introduce a confidence-aware VAE module that dynamically perturbs and regularizes $\tau_0$ based on its confidence score $s$. The latent feature of the trajectory is sampled from $\mathcal{N}(\mu, \sigma^2)$, where the variance $\sigma^2= \alpha (1 - s)$ and $\alpha$ as a scaling factor. This adaptive noise injection focuses refinement on regions of high uncertainty, improving both efficiency and targeting.

The encoded trajectory features are then fused with the granular driving command embedding through multi-head cross-attention, enabling dynamic alignment of latent maneuver representations with precise control objectives. A lightweight pre-trained Scorer module, introduced in the prior work~\cite{jiang2025irl} and trained on NAVSIM v2~\cite{dauner2024navsim}, selects the highest-scoring candidate trajectory $\tau^*$ by predicting the Extended Predictive Driver Model Score (EPDMS). This score assesses trajectories based on safety, efficiency, and comfort criteria, enabling simulator-free evaluation.

The transformer-based trajectory head refines candidate trajectories by leveraging geometric constraints, and latent distribution regularization. Specifically, the refinement loss $\mathcal{L}_{\text{refine}}$ is a weighted combination of an L1 regression term for geometric alignment, and a KL divergence constraint from the confidence-aware VAE:
\vspace{-2pt}
\begin{equation}
    \mathcal{L}_{\text{refine}} = \lambda_{\text{reg}} \cdot \mathcal{L}_{1}(\tau^{*(i)}, \tau_{\text{gt}}) + \lambda_{\text{kl}} \cdot D_{\text{KL}}\left( q(\mathbf{z}|\tau^{*(i)}) \parallel p(\mathbf{z}) \right),
\end{equation}
where $\mathcal{L}_{1}$ is the L1 loss deviation from ground truth trajectory $\tau_{\text{gt}}$, $p(\mathbf{z})$ is a standard Gaussian prior, and $q(\mathbf{z}|\tau^{*(i)})$ is the latent distribution for the $i$-th candidate of trajectory $\tau^{*}$, promoting safe and high-scoring trajectories.

Leveraging confidence-aware regularization, semantics-guided optimization, and multi-criteria scoring, our multi-stage planner progressively refines $\tau_0$ into a dynamically feasible and contextually appropriate trajectory. The hierarchical architecture ensures both high-level interpretability and low-level control precision: the Spatio-Temporal VLA generates semantic proposals and coarse trajectories, while the subsequent refinement stages handle geometric optimization.
\vspace{-5pt}
\subsection{Training Strategy}
\vspace{-2pt}
\label{sec:training} 
Our training process consists of three sequential stages, as illustrated in Fig~\ref{fig:pipeline}: First, we pre-train the dynamic token sparser. Next, we jointly optimize the sparser, LLM, and text encoder to enable reasoning of granular driving commands, coarse trajectories, and confidence scores. Finally, we train the command encoder and the transformer-based hierarchical trajectory head by minimizing $\mathcal{L}_{\text{refine}}$. The visual question-answering (VQA) data used in these stages are generated from ground truth trajectories, as described in Sec.~\ref{subsec:cot}.
\section{Experiments}\label{exp}
To validate the effectiveness of our proposed framework, we conduct comprehensive evaluations on the established NAVSIM v2 benchmark~\cite{dauner2024navsim}, comparing against a range of state-of-the-art methods. We further perform detailed ablation studies to isolate the contribution of each key component within our architecture.
\subsection{Experimental Setup and Metrics}
\begin{table*}[htbp!]
  \centering
  \setlength{\tabcolsep}{0.1cm} %
  \caption{Evaluation on NAVSIM v2 Navtest Benchmark for Planning oriented E2E Autonomous Driving.}
  \vspace{-2pt}
  \begin{tabular}{@{}l|c|c|ccccccccc|l@{}}
    \toprule
    \textbf{Method} & \textbf{VLM} & \textbf{VLA} & \textbf{NC $\uparrow$} & \textbf{DAC $\uparrow$} & \textbf{DDC $\uparrow$} & \textbf{TL $\uparrow$} & \textbf{EP $\uparrow$} & \textbf{TTC $\uparrow$} & \textbf{LK $\uparrow$} & \textbf{HC $\uparrow$} & \textbf{EC $\uparrow$} & \textbf{EPDMS $\uparrow$}\\
    \midrule
    Human Agent &\ding{55} &\ding{55} &100 &100 &99.8 &100 &87.4 &100 &100 &98.1 &90.1 &90.3 \\
    \midrule
    Transfuser\cite{chitta2022transfuser} &\ding{55} &\ding{55} &96.9 &89.9 &97.8 &99.7 &87.1 &95.4 &92.7 &98.3 &87.2 &76.7\\
    VADv2\cite{chen2024vadv2} &\ding{55} &\ding{55} &97.3 &91.7 &77.6 &92.7 &100 &99.9 &98.2 &66.0 &97.4 &76.6\\
    HydraMDP++\cite{lihydra} &\ding{55} &\ding{55} &97.2 &97.5 &99.4 &99.6 &83.1 &96.5 &94.4 &98.2 &70.9 &81.4\\
    GTRS-Dense\cite{li2025generalized} &\ding{55} &\ding{55} &97.6 &98.5 &99.5 &99.9 &89.5 &97.2 &96.8 &97.2 &57.2 &84.0\\
    GTRS-Dence w/ SimScale\cite{tian2025simscale} &\ding{55} &\ding{55} &98.4 &98.8 &99.4 &99.9 &87.9 &98.1 &96.4 &97.6 &58.8 &84.6\\
    DiffusionDrive\cite{liao2025diffusiondrive} &\ding{55} &\ding{55} &98.0 &96.0 &99.5 &99.8 &87.7 &97.1 &97.2 &98.3 &87.6 &84.3\\
    ZTRS\cite{li2025ztrs} &\ding{55} &\ding{55} &98.2 &99.1 &99.7 &99.8 &86.9 &97.5 &96.6 &98.2 &78.2 &86.2\\
    DriveSuprim\cite{yao2025drivesuprim} &\ding{55} &\ding{55} &98.4 &98.6 &99.6 &99.8 &90.5 &97.8 &97.0 &98.3 &78.6 &87.1\\
    Senna-E2E\cite{jiang2024senna}  &\ding{51} &\ding{55} &98.4 &97.4 &99.6 &99.8 &83.8 &97.9 &97.3 &98.3 &85.2 &84.8\\
    \textbf{HiST-E2E (Ours)} &\ding{51} &\ding{55} &98.6 &97.8 &99.6 &99.8 &83.9 &98.1 &97.3 &98.3 &85.1 &\textbf{85.4{\color{Green}($+1.3\%$)}}\\
    \textbf{HiST-VLA (Ours)} &\ding{51} &\ding{51} &99.6 &99.1 &99.7 &99.9 &89.2 &99.4 &98.9 &98.4 &66.0 &\textbf{88.6}\\
    \bottomrule
  \end{tabular}
  \label{tab:vlm_exp}
\end{table*}
\vspace{-2pt}
\begin{table*}[htbp!]
  \centering
  \setlength{\tabcolsep}{0.1cm} 
  \vspace{-2pt}
  \caption{Evaluation on NAVSIM-V2 Navhard Benchmark for Planning oriented E2E Autonomous Driving.}
  \vspace{-2pt}
  \centering
  \begin{tabular}{l|c|c|l| l l l l l l l l l |l}
    \toprule
    \textbf{Method} 
    & \textbf{VLM}
    & \textbf{VLA}
    & \textbf{Stage}
    & $\textbf{NC$\uparrow$}$
    & $\textbf{DAC$\uparrow$}$
    & $\textbf{DDC$\uparrow$}$
    & $\textbf{TLC$\uparrow$}$
    & $\textbf{EP$\uparrow$}$
    & $\textbf{TTC$\uparrow$}$
    & $\textbf{LK$\uparrow$}$ 
    & $\textbf{HC$\uparrow$}$ 
    & $\textbf{EC$\uparrow$}$ 
    & $\textbf{EPDMS$\uparrow$}$  \\
    \midrule
    
    PDM-Closed~\cite{dauner2023parting}  & \ding{55} & \ding{55}&   \makecell{Stage 1 \\ Stage 2} & \makecell{94.4 \\ 88.1} & \makecell{98.8 \\ 90.6} & \makecell{100 \\ 96.3} & \makecell{99.5 \\ 98.5} & \makecell{100 \\ 100} & \makecell{93.5 \\ 83.1 } & \makecell{99.3 \\ 73.7} & \makecell{87.7 \\ 91.5} & \makecell{36.0 \\ 25.4} & 51.3    \\
    \midrule
    Transfuser~\cite{chitta2022transfuser} & \ding{55} & \ding{55} &  \makecell{Stage 1 \\ Stage 2} &
    \makecell{96.2 \\ 77.7} & \makecell{79.5 \\ 70.2} & \makecell{99.1 \\ 84.2} & \makecell{99.5 \\ 98.0} & \makecell{84.1 \\ 85.1} & \makecell{95.1 \\ 75.6} & \makecell{94.2 \\ 45.4} & \makecell{97.5 \\ 95.7} & \makecell{79.1 \\ 75.9} & 23.1    \\
    \midrule

    DiffusionDrive~\cite{liao2025diffusiondrive} & \ding{55} & \ding{55} & \makecell{Stage 1 \\ Stage 2}  &
    \makecell{95.9 \\ 79.5} & \makecell{84.0 \\ 72.8} & \makecell{98.6 \\ 84.1} & \makecell{99.8 \\ 98.4} & \makecell{84.4 \\ 87.5} & \makecell{96.0 \\ 76.2} & \makecell{95.1 \\ 46.6} & \makecell{97.6 \\ 96.1} & \makecell{71.1 \\ 62.4} &  26.0  \\
    \midrule

    \makecell[l]{Senna-E2E\cite{jiang2024senna}}  & \ding{51} & \ding{55}&   \makecell{Stage 1 \\ Stage 2}  &
    \makecell{95.6 \\ 78.6} & \makecell{86.0 \\ 74.8} & \makecell{98.9 \\ 84.8} & \makecell{99.6 \\ 98.2} & \makecell{83.9 \\ 88.2} & \makecell{95.1 \\ 75.7} & \makecell{95.3 \\ 46.9} & \makecell{97.6 \\ 96.0} & \makecell{75.6 \\ 65.8} & 27.2    \\
    \midrule
    
    ZTRS~\cite{li2025ztrs} & \ding{55} & \ding{55} & \makecell{Stage 1 \\ Stage 2}  &
    \makecell{98.9 \\ 91.1} & \makecell{97.6 \\ 90.4} & \makecell{100 \\ 95.8} & \makecell{100 \\ 99.0} & \makecell{66.7 \\ 63.6} & \makecell{98.9 \\ 89.8} & \makecell{96.2 \\ 60.4} & \makecell{96.7 \\ 97.6} & \makecell{44 \\ 66.1} &  48.1  \\
    \midrule

    GTRS-Dence w/ SimScale\cite{tian2025simscale} & \ding{55} & \ding{55} & \makecell{Stage 1 \\ Stage 2}  &
    \makecell{99.1 \\ 92.3} & \makecell{98.2 \\ 93.8} & \makecell{99.8 \\ 94.9} & \makecell{100 \\ 99.3} & \makecell{71.9 \\ 75.4} & \makecell{99.3 \\ 90.7} & \makecell{95.6 \\ 60.3} & \makecell{93.8 \\ 95.9} & \makecell{28.0 \\ 37.4} &  47.2  \\
    \midrule
    
    GTRS-E\cite{li2025generalized} & \ding{55} & \ding{55} & \makecell{Stage 1 \\ Stage 2}  &
    \makecell{98.9 \\ 92.3} & \makecell{99.3 \\ 93.3} & \makecell{99.8 \\ 94.6} & \makecell{99.8 \\ 99.2} & \makecell{75.2 \\ 73.1} & \makecell{98.4 \\ 91.2} & \makecell{96.0 \\ 53.9} & \makecell{97.6 \\ 96.7} & \makecell{51.6 \\ 56.8} &  49.4  \\
    \midrule

    \makecell[l]{\textbf{HiST-E2E (Ours)}}  & \ding{51} & \ding{55} & \makecell{Stage 1 \\ Stage 2}  &
    \makecell{97.3 \\ 84.5} & \makecell{89.6 \\ 78.2} & \makecell{99.4 \\ 91.0} & \makecell{99.6 \\ 98.8} & \makecell{78.7 \\ 78.7} & \makecell{97.6 \\ 81.7} & \makecell{95.1 \\ 50.9} & \makecell{97.8 \\ 98.7} & \makecell{72.9 \\ 75.2} & \textbf{34.2{\color{Green}($+31.5\%$)}}    \\
    \midrule
    
    \makecell[l]{\textbf{HiST-VLA (Ours)}}  & \ding{51} & \ding{51}&   \makecell{Stage 1 \\ Stage 2}  &
    \makecell{99.8 \\ 86.4} & \makecell{98.7 \\ 83.7} & \makecell{99.7 \\ 89.9} & \makecell{99.6 \\ 98.5} & \makecell{87.3  \\ 87.5} & \makecell{99.6 \\ 83.1} & \makecell{99.6 \\ 56.7} & \makecell{97.8  \\ 96.3} & \makecell{73.3 \\ 60.9} &  \textbf{50.9}  \\

  \bottomrule
 \end{tabular}
\label{table:result_navhard}
\end{table*}

Our analysis uses two NAVSIM v2 benchmarks: the Navtest set for general open-loop planning, and the more challenging Navhard set, designed to evaluate robustness under pseudo closed-loop conditions through synthetic perturbations. Performance is rigorously evaluated using the EPDMS, a comprehensive score that aggregates nine critical sub-metrics: No At-Fault Collision (NC), Drivable Area Compliance (DAC), Driving Direction Compliance (DDC), Traffic Light Compliance (TLC), Ego Progress (EP), Time to Collision (TTC), Lane Keeping (LK), History Comfort (HC), and Extended Comfort (EC). The EPDMS score provides a holistic measure of an autonomous driving system's safety, efficiency, and comfort.

Our implementation builds upon LLaVA-v1.5-7B~\cite{liu2024improved} as the base VLM, which comprises 7 billion parameters, integrates ViT-L/14 from CLIP as the vison encoder, MLPs for vision-language adapter, and Vicuna-7B as the LLM. We enhance this architecture by incorporating depth-aware spatial encoding and replacing the standard adapter with our proposed dynamic token sparsification module, as detailed in Sections~\ref{subsec:Spatially_Aware_Visual_Encoding} and~\ref{subsec:dynamic_token_sparser}.

Our model is first pre-trained for 1 epoch on the NAVSIM trainval split using 8 × NVIDIA H20 GPUs with a batch size of 32 and a learning rate of \(1 \times 10^{-3}\). Subsequently, stage-1 training is performed under the same hardware setup for 1 epoch on trainval, with a batch size of 8 and a learning rate of \(1 \times 10^{-5}\). Finally, stage-2 training is conducted on the navtrain split for 30 epochs using 8 × NVIDIA A800 GPUs, a batch size of 2, and a learning rate of \(1 \times 10^{-5}\).

We evaluate two variants of our proposed model: \textbf{HiST-E2E} uses a ResNet-34~\cite{resnet} backbone to extract visual features, HiST-VLA to generate granular driving commands only, and then feeds these features to an external motion planner (DiffusionDrive~\cite{liao2025diffusiondrive}) to produce the final trajectory; and \textbf{HiST-VLA}, our fully integrated ``pure" VLA model that directly generates trajectories without relying on any external E2E modules, operating as a single unified system.
\subsection{Quantitative Results}
\subsubsection{Performance on Navtest Benchmark}
Table~\ref{tab:vlm_exp} summarizes the experimental results on the NAVSIM v2 Navtest benchmark. Notably, our HiST-VLA model establishes a new state-of-the-art with an EPDMS score of 88.6, significantly outperforming established baselines such as VADv2 (76.6), Transfuser (76.7), and prior SOTA methods including DriveSuprim (87.1), the diffusion-based DiffusionDrive (84.3), GTRS-Dence w/ SimScale (84.6) and ZTRS (86.2). Remarkably, this score closely approaches the human expert performance of 90.3, demonstrating the model’s human-like decision-making capabilities. Such robust performance is anchored by exceptional results in safety-critical metrics, specifically NC (99.6), DAC (99.1), DDC (99.7), and LK (98.9), proving the effectiveness of our hierarchical design in ensuring driving safety. Crucially, the HiST-E2E variant, which fuses the driving command only from HiST-VLA and uses an external E2E planner (DiffusionDrive~\cite{liao2025diffusiondrive}) to produce the final trajectory, also attains a competitive EPDMS score of 85.4, surpassing the DiffusionDrive baseline (84.3). This result validates that our spatio-temporal reasoning mechanism provides substantial performance gains even when integrated with external planners, highlighting its generalizability. Collectively, these results confirm the superiority of our proposed framework.

\subsubsection{Performance on Navhard Benchmark}
Table~\ref{table:result_navhard} presents the evaluation results on the rigorous Navhard benchmark, which is to test robustness in pseudo closed-loop scenarios. Our HiST-VLA model achieves a competitive EPDMS score of 50.9, substantially outperforming established baselines such as DiffusionDrive (26.0) and Transfuser (23.1), while exhibiting superior performance compared to GTRS-E (49.4). Furthermore, the HiST-E2E variant attains an EPDMS of 34.2, representing a remarkable 31.5\% relative improvement over the DiffusionDrive planner baseline (26.0). This substantial gain validates that our spatio-temporal reasoning mechanism effectively enhances trajectory quality even when integrated with external planners. Collectively, these results demonstrate the robustness and generalizability of the HiST-VLA architecture across both open-loop (Navtest) and pseudo closed-loop (Navhard) settings.
\subsection{Ablation Studies}
To evaluate the contribution of each proposed component, we conduct three ablation studies on Navtest benchmark.

\subsubsection{Component Analysis of Spatio-Temporal VLA}
\begin{table}[htbp!]
  \centering
  \vspace{-5pt}
  \caption{Ablation study of Spatio-Temporal VLA modules on Navtest. }
  \vspace{-5pt}
  \setlength{\tabcolsep}{0.1cm} 
  \begin{tabular}{@{}c|c|c|c@{}}
    \toprule
    \makecell{\textbf{Spatial} \\ \textbf{Encoding}} & \makecell{\textbf{Temporal} \\ \textbf{Prompting}} & \makecell{\textbf{Granular} \\ \textbf{ Meta-Action}} & \textbf{EPDMS $\uparrow$}\\
    \midrule
    \ding{55} &\ding{55} &\ding{55}  &77.2\\
    \ding{55} &\ding{51} &\ding{51}  &85.0\\
    \ding{51} &\ding{55} &\ding{51}  &86.3\\ 
    \ding{51} &\ding{51} &\ding{55}  &83.3\\ 
    \midrule
    \ding{51} &\ding{51} &\ding{51} &\textbf{88.6}\\
    \bottomrule
  \end{tabular}
  \label{tab:st_vla_exp}
\end{table}
Table~\ref{tab:st_vla_exp} presents an ablation study evaluating the contribution of each component in VLA model. The full HiST-VLA model achieves the highest EPDMS of 88.6, showing substantial gains over the baseline of 77.2, where we simply integrate LLaVA-7B model with the proposed hierarchical planner. Removing spatial encoding results in a score of 85.0, while omitting temporal prompting causes a drop to 86.3. The token sparsification module contributes modestly to the overall performance, with its absence reducing the score to 87.9. Most substantially, ablating the Granular Meta-Action module leads to a pronounced decrease to 83.3, underscoring its critical role. These results demonstrate that all components contribute synergistically to the framework's effectiveness, with the complete system achieving a remarkable 11.4-point improvement over the baseline.

\subsubsection{Efficiency Analysis of Dynamic Token Sparser Module}
\begin{table}[htbp!]
    \scriptsize
    \centering
    \vspace{-5pt}
    \caption{Efficiency Analysis of Spatio-Temporal VLA on Navtest.}
    \vspace{-5pt}
    \label{tab:flops_eval}
    \begin{tabular}{c | l l }
        \toprule
         \textbf{Dynamic Token Sparser}             &  \textbf{GFLOPs} $\downarrow$ & \textbf{EPDMS} $\uparrow$ \\
        \midrule
        w/o & 9,105  &  87.9 \\
        w/  &  \textbf{7,630{\color{Green}($-23.7\%$)}} & \textbf{88.6}  \\
        \bottomrule 
    \end{tabular}
\end{table}
Table~\ref{tab:flops_eval} demonstrates the pivotal role of the dynamic token sparser in balancing computational efficiency and planning performance. Without the sparser, the model operates at a high computational cost of 9,105 GFLOPs, achieving an EPDMS of 87.9. By integrating the sparser with a fusion rate of 0.8, which effectively merges 20\% of redundant vision tokens, the model eliminates visual noise and significantly reduces the computational load to 7,630 GFLOPs while simultaneously boosting the EPDMS to 88.6.

\subsubsection{Module Analysis of Hierarchical Planner}
\begin{table}[htbp!]
  \centering
  \vspace{-5pt}
  \caption{Ablation study of Hierarchical planner on Navtest. }
  \vspace{-5pt}
  \setlength{\tabcolsep}{0.1cm} 
  \begin{tabular}{@{}c|c|c|c@{}}
    \toprule
    \makecell{\textbf{HiST-VLA} \\ \textbf{Coarse Traj.}} & \makecell{\textbf{Semantics} \\ \textbf{Alignment}} & \makecell{\textbf{Regularization} \\ \textbf{\& Scoring}} &\textbf{EPDMS $\uparrow$}\\
    \midrule
    \ding{51} &\ding{55} &\ding{55}  &73.1\\
    \ding{51} &\ding{51} &\ding{55}  &84.3\\
    \ding{51} &\ding{55} &\ding{51}  &85.0\\
    \midrule
    \ding{51} &\ding{51} &\ding{51}  &\textbf{88.6}\\
    \bottomrule
  \end{tabular}
  \label{tab:hi_vla_exp}
\end{table}
Table~\ref{tab:hi_vla_exp} evaluates the hierarchical modules within the planner of our HiST-VLA framework. The baseline model, which outputs only a coarse trajectory without any planner, achieves an EPDMS score of 73.1. Incorporating the semantics-aligned granular driving command encoding module by a transformer-based planner leads to a significant improvement, raising the score to 84.3. When the confidence-aware regularization and the multi-criteria scorer is added to refine the coarse trajectory, performance increases to 85.0 EPDMS score, highlighting its essential role in enhancing safety and compliance through trajectory evaluation and correction.

These results demonstrate that the proposed hierarchical components—individually and collectively—contribute substantially to the overall performance, establishing a new state-of-the-art in VLA autonomous driving systems.
\subsection{Trajecotory Visualization of Hierarchical Planner}
\label{subsec:vis_refinement}
\begin{figure}[htbp]
\centerline{\includegraphics[width=3.5in]{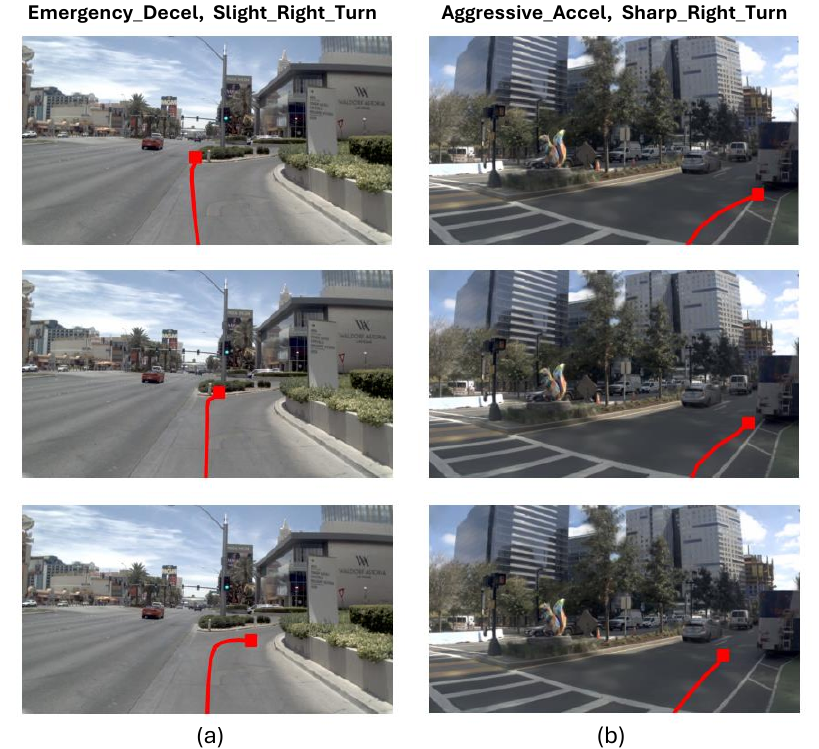}}
\vspace{-8pt}
\caption{Trajectory visualization on Navhard across: (a) stage 1 real-world and (b) stage 2 synthetic scenes. Driving commands are shown above each scenario. Rows depict, from top to bottom: VLA coarse trajectory, semantics-aligned trajectory, and HiST-VLA trajectory.}
\vspace{-5pt}
\label{fig:vis}
\end{figure}
Figure~\ref{fig:vis} illustrates the hierarchical refinement of trajectories across two distinct scenarios: (a) stage 1 real-world and (b) stage 2 synthetic environments. In scenario (a), the coarse trajectory exhibits ambiguous intent, partially intruding into green spaces. In contrast, the trajectory refined by the confidence-aware regularization and scorer module is both smoother and more tightly aligned with the drivable area. In scenario (b), both the coarse and semantics-aligned trajectories infringe upon parked vehicles to the right. The final hierarchically-refined trajectory, however, successfully avoids these obstacles, demonstrating improved adaptability and safety awareness. These visual comparisons highlight the progressive enhancement achieved through our refinement pipeline. Each stage effectively addresses limitations present in the previous one, resulting in a robust and human-like motion planning outcome.
\section{Conclusion}
This paper introduces HiST-VLA, a novel VLA framework designed to elevate autonomous driving systems through integrated 3D geometric reasoning and hierarchical trajectory refinement. By eliminating reliance on external E2E modules and expensive annotations, our approach significantly enhances spatial awareness and motion planning fidelity. The core of our architecture lies in two innovations: a spatio-temporally grounded reasoning module featuring dynamic token sparsification, and a hierarchical planner that aligns language commands with trajectory generation through confidence-aware regularization and multi-criteria scoring. Extensive experiments on NAVSIM v2 show that HiST-VLA achieves state-of-the-art performance on Navtest open-loop benchmark, with an EPDMS score of 88.6, markedly surpassing existing methods. Furthermore, comprehensive ablation studies validate the critical contribution of each individual component to the overall system efficacy.
\vspace{-10pt}
\bibliographystyle{IEEEtran}
\bibliography{Bibliography}

\end{document}